\documentclass[a4paper,10pt]{article}
\usepackage[english]{babel}
\usepackage[utf8]{inputenc}
\usepackage{multicol}
\usepackage{tablefootnote}
\usepackage[ruled,vlined]{algorithm2e}
\SetKwInOut{Parameter}{Parameters}
\usepackage{graphicx}

\title{Exploring Novel Quality Diversity Methods For Generalization in Reinforcement Learning}
\author{
Brad Windsor \thanks{Equal contribution} \thanks{New York University} \thanks{Thanks to Julian Togelius,
Lisa Soros, and 
Akhmed Khalifa for their oversight of this work}
\and
Brandon O'Shea \footnotemark[1] \footnotemark[2]
\and
Mengxi Wu \footnotemark[1] \footnotemark[2]

}
\begin{document}

\date{}
\maketitle

\begin{abstract}
    The Reinforcement Learning field is strong on achievements and weak on reapplication; a computer playing GO at a super-human level is still terrible at Tic-Tac-Toe. This paper asks whether the method of training networks improves their generalization. Specifically we explore core quality diversity algorithms, compare against two recent algorithms, and propose a new algorithm to deal with shortcomings in existing methods. Although results of these methods are well below the performance hoped for, our work raises important points about the choice of behavior criterion in quality diversity, the interaction of differential and evolutionary training methods, and the role of offline reinforcement learning and randomized learning in evolutionary search. 
\end{abstract}

\begin{multicols}{2}

\medskip

\section*{Introduction}
\subsection*{Reinforcement Learning and Generalization}

Reinforcement Learning \cite{10.5555/551283} is a learning paradigm where an agent interacts with its environment to maximize a reward. General algorithms developed within this framework have potential application across autonomous vehicles, digital assistants, healthcare treatment, and others. To study and develop new algorithms, researchers often rely on toy problems such as video games.

While full of promising results, one challenge across Reinforcement Learning applications is that results on one task rarely generalize well to similar tasks. A neural network which can play a game will fail when the level is cropped or written in its mirror image \cite{ye2020rotation}, and fail on games with similar mechanics. This raises questions about the value of the information learned; have networks understood gameplay strategies or have they merely memorized sequences of steps in response to a set of pixels?

While the Actor-Critic learning algorithm has been shown to be effective in mastering our toy problem \cite{justesen2018illuminating}, this project focused on whether evolving multiple solutions through enhanced Quality Diversity methods would improve the generalizability of neural network solutions. We explore whether evolution and a variety of interesting solutions on previous levels lead to a higher likelihood of solving future levels.

\subsubsection*{Toy Problem}
We focus on Zelda as the example problem. Zelda is a sparse-reward problem with many-step solutions; although getting the key has a low reward signal, it is essential for completion.

\begin{figure}
\includegraphics[scale=0.35]{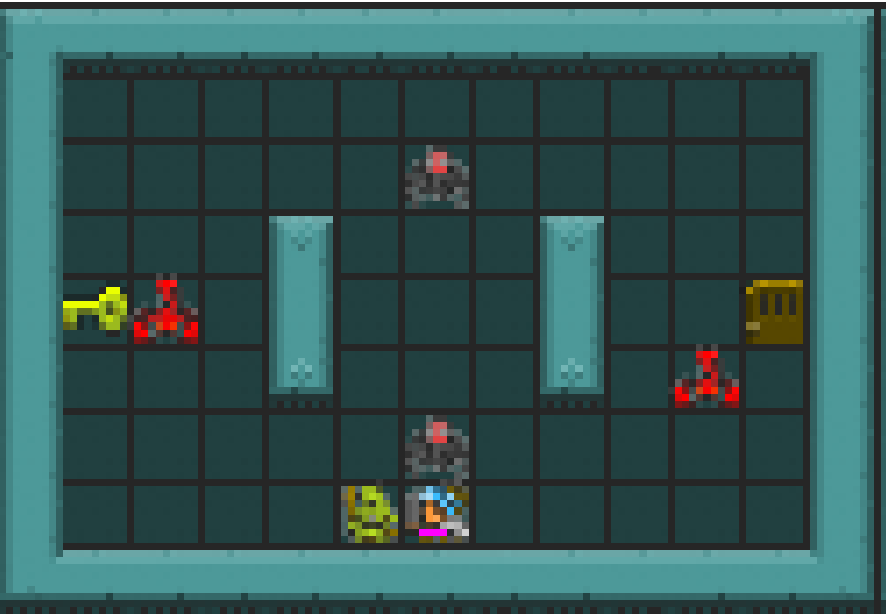}
\caption{Fig 1. A Sample Zelda Level. The agent must get the key, avoid or kill monsters, and reach the door. }
\end{figure}

\subsection*{Overview of Quality Diversity Algorithms}
Quality Diversity refers to a category of algorithms which, instead of seeking to optimize for a problem directly, evolve the best-possible result in one or more categories of solutions. The metric for categorizing the solution is called a \textit{behavior criterion} (BC), which can be any way of characterizing an agent's features, actions, or results. Quality Diversity methods are inspired by natural evolution and include algorithms such as multi-BC novelty search, and multi-BC Map Elites.

\textbf{Map Elites} \newline
Map Elites \cite{mouret2015illuminating} is a search algorithm where a high dimensional feature space is mapped to a lower set of dimensions, and the best performing agent is tracked in each grid cell of the lower-dimensional space. In this paper, we use the levels of Zelda to characterize the solution space, such that millions of possible actions and results were reduced to a win and loss on each of 10 Zelda levels, 1024 possible agents.

Several experiments were done on the vanilla algorithm:
\begin{enumerate}
    \item Varying the standard deviation parameter for sampling new networks
    \item A character-based tile map instead of a pixel-based view of the game
    \item Augmenting the reward function to add a point whenever a new square was visited
    \item Varying the map representation such that the best agent is tracked if it wins the key, or if it wins the key and the level. As the key is a prerequisite to opening the door, this is an intermediate level solution.
\end{enumerate}

\textbf{Covariance Matrix Adaptation Map Elites (CMAME)} \newline
Covariance Matrix Adaptation Map Elites \cite{Fontaine_2020} is an application of Covariance Matrix Adaptation Evolution Strategy (CMA-ES) \cite{hansen2016cma} to the Map Elites framework. Instead of randomly generating candidate solutions, as with Vanilla Map Elites, CMA-ES keeps a covariance matrix between all features and uses it to generate solutions that more efficiently explore the feature space. 

While the original CMAME uses the parameter feature space to designate map cells, our implementation of CMAME follows that of Map Elites code in using level wins as the map.

\textbf{Differential Map Elites (DME)} \newline
DME \cite{Togelius_2020} is a combination of map elites with differential evolution, an optimization algorithm which randomly distributes agents over the search space, then uses the difference in candidate feature vectors to and performances to propose new candidate performances. It is like CMAME in that it uses calculated metaparameters to more intelligently propose solutions, but differs in the candidate generation algorithm.

\end{multicols}
\begin{algorithm}[H]
\SetAlgoLined
  \KwIn{$ExploreMap$, $FollowMap$, $CountEvaluated$}
  \Parameter{$Startup$, $ExploreRatio$}
  \KwOut{An agent to be evaluated}    

  \uIf{$CountEvaluated$ $<$ $Startup$ }{
    \Return $NewAgent(ExploreMap)$
   }
   \uElseIf{$Random()$ $<$ $ExploreRatio$}{
    \Return $NewAgent(ExploreMap)$
  }
  \Else{
   \Return $NewAgent(FollowMap)$
  }
 \caption{EFME: Get Next Agent For Evaluation}
\end{algorithm}

\begin{algorithm}[H]
\SetAlgoLined
  \KwIn{$ExploreMap$, $FollowMap$, $Agent$, $Feature$, $Score$, $Timesteps$}
  \uIf{$Score$ $>$ $ExploreMap.GetScore(Feature)$ }{
    $ExploreMap$.$SetAgent(Agent, Feature, Score)$
   }
  \uIf{$Timesteps$ $<$ $FollowMap.GetTimesteps(Feature)$ }{
    $FollowMap$.$SetAgent(Agent, Feature, Score)$
   }
 \caption{EFME: Update Explore and Follow Maps}
\end{algorithm}
\begin{multicols}{2}

\textbf{Explorer-Follower Map Elites (EFME)} \newline
EFME is a new algorithm inspired by the observation that both ME and CMA-ME runs frequently converged when more timesteps per level are allowed. EFME seeks to evolve efficient, lower stepcount solutions from inefficient, randomized solutions evolved at a higher step count. Although Zelda levels can be solved in 200 steps or fewer, we note one Atari benchmark done with 20,000 timesteps \cite{chrabaszcz2018basics}, so the practice of allowing agents extra steps to solve levels is not uncommon.

Algorithms 1 and 2 give a detailed explanation of EFME. Two maps are maintained, an $ExploreMap$ which maps $Feature$ to $(Agent, Score)$, and a $FollowMap$ which maps $Feature$ to $(Agent, TimeSteps)$. The $ExploreMap$ updates when a higher-scoring agent is found, while the $FollowMap$ updates when an agent solves the same levels more efficiently. The algorithm samples from the $ExploreMap$ with frequency $ExploreRatio$. An evaluated agent may update either or both maps. 

Both EFME is in the spirit of of Go-Explore \cite{ecoffet2019goexplore} in that they first solve a problem then solve it more efficiently. However, DPRL differs in (1) a different explore method, (2) treating the entire level as one trajectory, (3) using the reward signal to highlight a variety of especially important states, and (4) the use of (3) for pretraining, including backpropagation-based pretraining.

\begin{figure}[h]
    \centering
    \includegraphics[scale=0.55]{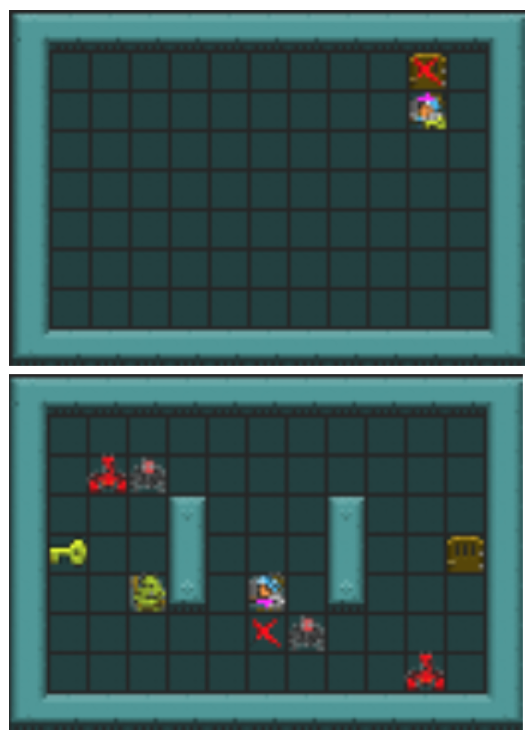}
\caption{Fig 2. Sample puzzles. Top: move to win. Bottom: attack to score. }
\end{figure}

\section*{Experimental Setup}
All code is available online \footnote{https://github.com/bwindsor22/quality-diversity-rl}.

\subsection*{QD Algorithms}
For vanilla map elites, new candidate solutions are generated by changing each parameter with a probability of 0.7 and setting each value by sampling from a distribution with a variance of 0.03, the same settings as in \cite{lazarou} (0.02 is used in \cite{salimans2017evolution}).

This project included the first Python write of CMAME, a port from the author's C\# code \footnote{https://github.com/tehqin/QualDivBenchmark}, and  leveraging Hansen's library for CMA-ES. \footnote{https://github.com/CMA-ES/pycma} 

\subsection*{Evaluation Framework}
Experiments were performed on the General Video Game playing framework \cite{perezliebana2018general}. We used a deterministic Zelda implementation for more reproducible results, and included five easier Zelda training levels from \cite{ye2020rotation}.

Initial experiments were performed on a single 32-core host with an object pool of cached GVG-AI levels. Experiments with greater than 20K map elites iterations were performed at the NYU High Performance Computing (HPC) facility, where a parent process ran the core Map Elites algorithm, and several child processes evaluated candidate networks on levels. Typically 200 children were used, and a 400,000 iteration run is approximately 3 days of elapsed time, depending on the algorithm. Other optimizations include:
\begin{enumerate}
	\item An option to run a faster unix-friendly version of GVG-AI \footnote{https://github.com/Bam4d/GVGAI\_GYM}
	\item A caching level maker, which improved run time by resetting evaluation environments when possible
	\item Several utilities to save and replay games or identify game states
\end{enumerate}

Unless otherwise noted, experiments were performed with a 7K parameter network of three convolutional layers and one dense layer.

\section*{Results}
In the results section, column titles are defined as follows:
\begin{itemize}
    \item \textit{Fitness Evaluations} refers to end to end runs of an agent on all 10 levels
    \item \textit{Agents in Map At Finish} refers to the count of different agents in a Map Elites implementation. A sample agent might be denoted as  1-0-1-0-0-0-0-0-0-0 if it won the first and third levels and no others. 1024 agents are possible for most trials, $2^{20}=1,048,576$ when the key win is treated separately.
    \item \textit{Most Levels Solved} refers to the highest count of levels solved by any individual agent. For 1-0-1-0-0-0-0-0-0-0, this number is 2.
\end{itemize}

\subsubsection*{Vanilla Map Elites and Modifications}
Tables \ref{vme} through \ref{vmetm} show evaluation for Vanilla Map Elites. Although long-running evaluations did give rise to 10 agents, no agent came close to solving all levels. There is no evidence that modifying the Gaussian noise standard deviation at onset meaningfully changes the result, which is in line with \cite{salimans2017evolution} who find that varying this parameter at runtime gives no difference.

Switching to a tile map view and including an explore reward did not change the highest count of levels solved. However, slightly relaxing the steps per level and adding partial map elites cells for gaining a key allowed the agent to solve levels far sooner (at ~80K vs ~400K iterations).  Including key cells did have a meaningful difference in the number of elites in the map, including 6 elites which acquired the key in a level other than the one solved. Using prior knowledge to identify intermediate points in level solving is highly beneficial in solving levels more quickly and in increasing the diversity of solutions tracked in the map elites map. 

\subsubsection*{DME and CMAME}
Our initial DME run was the longest of any to date, 871,000 iterations for almost a week runtime, but failed to solve more than one level. CMAME with a lower reset frequency was the most effective of the modified algorithms, though the performance of both was comparable to vanilla Map Elites.

\subsubsection*{EFME}
Explorer-Follower Map Elites was the only approach which seemed to fully populate the agent map and solve more than one level. A higher explore ratio solved more levels sooner, but a lower one led to a lower average step count per level (16,461 vs ~30,000). The follow map appears effective for evolving lower-step solutions, though none nears the human performance of ~200 steps per level.

One note in EFME is that the growth of agents in the map is nearly logarithmic $(R^2 = 0.89)$. To reach even 300 of the possible 1024 agents, the closest-fit logarithmic curve indicates that almost 50 million evaluations would be needed.

\begin{figure}[h]
\includegraphics[scale=0.45]{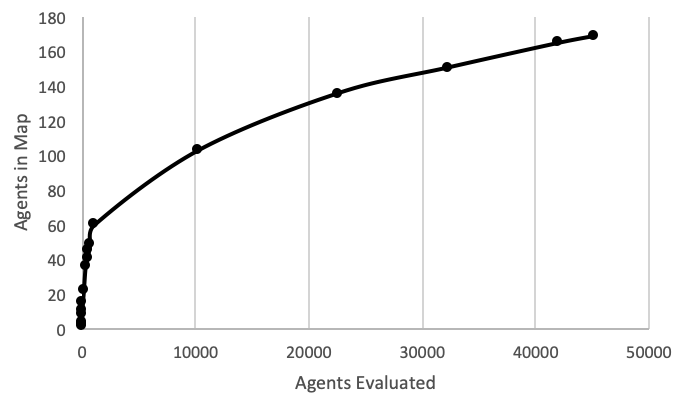}
\caption{Fig 3. Growth of Agents in the Explore Map}
\end{figure}

\end{multicols}
\begin{table}[]
\caption{Vanilla Map Elites Evaluations} \label{vme}
\begin{tabular}{|l|l|l|}
\hline
Fitness Evaluations & Agents in Map at Finish & Most Levels Solved \\ \hline
570,000             & 10                      & 2                  \\ \hline
200,000             & 9                       & 2                  \\ \hline
50,000              & 4                       & 2                  \\ \hline
\end{tabular}

\caption{Vanilla Map Elites at Varying Gaussian Noise Std Dev} \label{stddev}
\begin{tabular}{|l|l|l|l|}
\hline
Fitness Evaluations & Gaussian Noise Std Dev & Agents in Map at Finish & Most Levels Solved \\ \hline
400,000             & 0.5                    & 5                       & 2                  \\ \hline
100,000             & 0.1                    & 2                       & 1                  \\ \hline
\end{tabular}

\caption{Vanilla Map Elites with Tile Map and Explore Reward} \label{vmetm}
\begin{tabular}{|l|l|l|}
\hline
Fitness Evaluations & Agents in Map at Finish & Most Levels Solved \\ \hline
300,000             & 5                       & 1                \\ \hline
\end{tabular}

\caption{Vanilla Map Elites with Tile Map, Explore Reward, and Partial Key Levels} \label{vmetm}
\begin{tabular}{|l|l|l|l|}
\hline
Fitness Evaluations & Steps Per Level & Agents in Map at Finish \footnotemark & Most Levels Solved                      \\ \hline
74,000              & 2,000           & 38                       & 1                  \\ \hline
80,000              & 2,000           & 56                       & 1                 \\ \hline
\end{tabular}

\footnotetext[5]{Modified map. See item 4 in Map Elites section}

\caption{DME Results} \label{dme}
\begin{tabular}{|l|l|l|l|}
\hline
Fitness Evaluations & Agents in Map at Finish & Most Levels Solved & Note              \\ \hline
871,000             & 2                       & 1                  &                   \\ \hline
500,000             & 8                       & 1                  & Removed crossover \\ \hline
\end{tabular}

\caption{CMAME Results} \label{cmame}
\begin{tabular}{|l|l|l|l|}
\hline
Fitness Evaluations & Agents in Map at Finish & Most Levels Solved & Note                             \\ \hline
350,082             & 7                       & 2                  &                                  \\ \hline
400,000             & 3                       & 1                  & Higher reset freq \\ \hline
\end{tabular}

\caption{Explorer-Follower Results} \label{efr}
\begin{tabular}{|l|l|l|l|l|}
\hline
Fitness Evaluations & Steps Per Level & Agents in Map at Finish & Most Levels Solved & Explore Ratio \\ \hline
13,000              & 50,000          & 277                     & 7                  & 1             \\ \hline
45,000              & 50,000          & 166                     & 6                  & 0.67          \\ \hline
\end{tabular}
\end{table}

\begin{multicols}{2}

\section*{Conclusion}
This project evaluated a core map elites algorithm, the effect of three changes in environmental setup on the core algorithm, two advanced forms of the map elites algorithm and two novel map-elites inspired algorithms. 

Some of the key takeaways are:
\begin{itemize}
    \item Networks fail to take the lessons learned from one level and apply them to other levels. In the case of EFME, 99.8\% of evaluated agents were mutated from an agent which could already solve at least one level, but were unable to solve many others. EFME ends with hundreds of cases where a network can solve 3-5 levels but not others. 
    \item The choice of evolutionary training strategies may require more training time to memorize long sequences of steps as compared to backpropagation methods \cite{ye2020rotation}, and there is little discernible evidence of improved generalization.
    \item The characterization of agents by levels solved does not appear to work well with map-elites-based algorithms. We conclude that Map Elites cells must be more closely tied to the network parameters as in \cite{Fontaine_2020}, or to intermediate goals as in the partial key levels in this project or direction traveled in a maze \cite{10.3389/frobt.2016.00040}. Tracking good solutions for level 1 and trying them on level 2 is asking the evolved network to generalize, something they do poorly.
    \item In a sparse reward environment with multi-step solutions and a poorly-chosen behavior criterion, neither DME nor CMAME significantly outperform their vanilla ME counterpart.
    \item Evolution Strategies most easily solve video games when given high step counts which allow very randomized agents to act.
    \item There is some early evidence that highly randomized agents may still be useful in finding beneficial behavior for use in imitation learning, and that backpropagation methods can give a jump start to evolutionary ones.
\end{itemize}

\subsubsection*{Acknowledgements}
The authors wish to thank Shenglong Wang, Matt Fontaine, Michael Green, and Aaron Dharna for their assistance in experiments and valuable insight.

\medskip

\bibliographystyle{unsrt}
\bibliography{sample}

\end{multicols}
\end{document}